\documentclass{article}

\PassOptionsToPackage{numbers, compress}{natbib}

\usepackage[final]{neurips_2024}

\usepackage[utf8]{inputenc} 
\usepackage[T1]{fontenc}    
\usepackage{url}            
\usepackage{booktabs}       
\usepackage{amsfonts}       
\usepackage{nicefrac}       
\usepackage{microtype}      
\usepackage[numbers]{natbib}
\usepackage[dvipsnames,table,xcdraw]{xcolor}
\definecolor{cvprblue}{rgb}{0.21,0.49,0.74}
\usepackage{graphicx}
\usepackage{caption}
\usepackage{arydshln}
\usepackage{tikz}
\usepackage{wrapfig}
\usepackage{float}
\usepackage{multirow}
\usepackage{pifont}
\usepackage{makecell}
\usepackage{subfigure}
\usepackage[pagebackref,breaklinks,colorlinks,citecolor=cvprblue]{hyperref}

\newcommand{\tabref}[1]{Tab.~\ref{#1}}

\def\MyMthd{HumanOmni}

\newcommand{\tablestyle}[2]{\setlength{\tabcolsep}{#1}\renewcommand{\arraystretch}{#2}\centering\small}

\title{HumanOmni: A Large Vision-Speech Language Model for Human-Centric Video Understanding}

\author{
\textnormal{Jiaxing Zhao}$^{1\dag}$\footnotemark[1]\quad
\textnormal{Qize Yang}$^{1}$\footnotemark[1] \quad
\textnormal{Yixing Peng}$^{2,1}$\footnotemark[1] \quad
\textnormal{Detao Bai}$^{1}$\footnotemark[1] \quad
\textnormal{Shimin Yao}$^{1}$\footnotemark[1] \quad 
\textnormal{Boyuan Sun}$^{3,1}$ \quad \\
Xiang Chen$^{1}$\quad
Shenghao Fu$^{2,1}$\quad
Weixuan Chen$^1$\quad
Xihan Wei$^1$\quad 
Liefeng Bo$^1$\quad\\
{\normalsize{$^1$Tongyi Lab, Alibaba Group}} \quad
{\normalsize{$^2$ISEE, Sun Yat-sen University}} \quad
{\normalsize{$^3$VCIP, CS, Nankai University}}\\
{\tt{\small{zjx244036@alibaba-inc.com}}}\\
\textbf{\normalsize{\url{https://github.com/HumanMLLM/HumanOmni}}}\vspace{-3mm}}

\begin{document}

\maketitle

\footnotetext[1]{Equal contribution} 
\footnotetext[2]{Project Leader}

\begin{abstract}
  In human-centric scenes, the ability to simultaneously understand visual and auditory information is crucial. 
While recent omni models can process multiple modalities, they generally lack effectiveness in human-centric scenes due to the absence of large-scale, specialized datasets and non-targeted architectures. 
  In this work, we developed HumanOmni, the industry's first human-centric Omni-multimodal large language model. 
    We constructed a dataset containing over 2.4 million human-centric video clips with detailed captions and more than 14 million instructions, facilitating the understanding of diverse human-centric scenes. 
    HumanOmni includes three specialized branches for understanding different types of scenes.
    It adaptively fuses features from these branches based on user instructions, significantly enhancing visual understanding in scenes centered around individuals.
    Moreover, HumanOmni integrates audio features to ensure a comprehensive understanding of environments and individuals. 
    Our experiments validate HumanOmni's advanced capabilities in handling human-centric scenes across a variety of tasks, including emotion recognition, facial expression description, and action understanding.
  Our model will be open-sourced to facilitate further development and collaboration within both academia and industry.
\end{abstract}


    

\section{Introduction}
In the era of rapid digital and intelligent development, understanding human-centric scenes has become increasingly critical. These scenes extend beyond video chat~\cite{li2023videochat} to encompass education, healthcare, social interactions, and entertainment. In these human-centric scenes, vision and speech are typically present simultaneously. For certain tasks, both visual and auditory information provide significant benefits, such as in emotion recognition~\cite{chen2024motionllm, cheng2024emotion, liu_mafw_2022,zhao2023dferclip} and speaker-specific speech recognition. 
Speaker-specific speech recognition builds upon automatic speech recognition by incorporating additional description about the speaker.
We are currently defining this task and collecting such a dataset, with plans to release it in the next version of our work. 

Current methods predominantly focus on Vision-Language models~\cite{alayrac2022flamingo, dong2024internlm, li2023llava-med, li2024llava, liu2024llavanext, openai2023gpt4, gpt4v, geminiteam2024geminifamilyhighlycapable,   ye2023mplugowl, ye2023mplug, ye2023mplugowl2, zhu2023minigpt}, which effectively handle visual and textual information but generally lack the capability to process audio inputs. 
This limitation results in an incomplete understanding of scenes.
In recent years, some omni models~\cite{fang-etal-2024-llama-omni,fu2024vitaopensourceinteractiveomni, li2024baichuanomni, xie2024miniomnilanguagemodelshear,yang2025omni} have been proposed to address multiple modalities, including visual, auditory, and textual data. 
However, these models often emphasize generic scenes and lack targeted training for human-centric scenes. Additionally, they do not incorporate specialized model designs, leading to weaker performance in understanding such scenarios.

Moreover, there are specialized models designed specifically for specific tasks that have incorporated both audio and video inputs. These models have demonstrated significant effectiveness in their targeted applications. However, their specificity limits their generalizability. Due to their narrow focus and reliance on specific datasets or conditions, these specialized models perform poorly when applied to broader, more diverse human-centric scenes. 

 In this work, we present HumanOmni, a large vision-speech language
model for human-centric video unstanding.
The key feature of HumanOmni is its ability to simultaneously process vision and speech information in human-centric scenes. 
It achieves excellent performance in various human-centric scenes. 
Our contributions can be summarized in three key areas:

\begin{itemize}
    \item We have constructed a dataset containing over 2.4M human-centric video clips, providing rich and detailed information about individuals. 
    We provide over 14M instruction data for visual pretraining.  
    Additionally, we have manually annotated 50K video clips with more than 100K instructions related to emotion recognition, facial description, and speaker-specific speech recognition for visual fine-tuning and cross-modal interaction integration. 
    This comprehensive data enables our model to better understand individual characteristics and the human-centric scenes.
    \item We use three branches to handle face-related, body-related, and interaction-related scenes separately in HumanOmni. An instruction-driven fusion module then integrates features from these branches. HumanOmni dynamically adjusts its fusion weights based on input instructions, ensuring accurate responses across various scenes.
    \item HumanOmni can simultaneously understand vision and speech, allowing for a more comprehensive understanding of complex scenes. Our experiments show that HumanOmni achieves state-of-the-art performance on various tasks, outperforming existing Vision-Language models, Omni models, and even specialized proprietary models in their respective domains. Additionally, HumanOmni excels in audio-only tasks like automatic speech recognition, delivering results comparable to leading models in this field.
\end{itemize}

\section{Our Model}
\begin{figure*}
    \vspace{-5mm}
    \centering
    \includegraphics[width=1\linewidth]{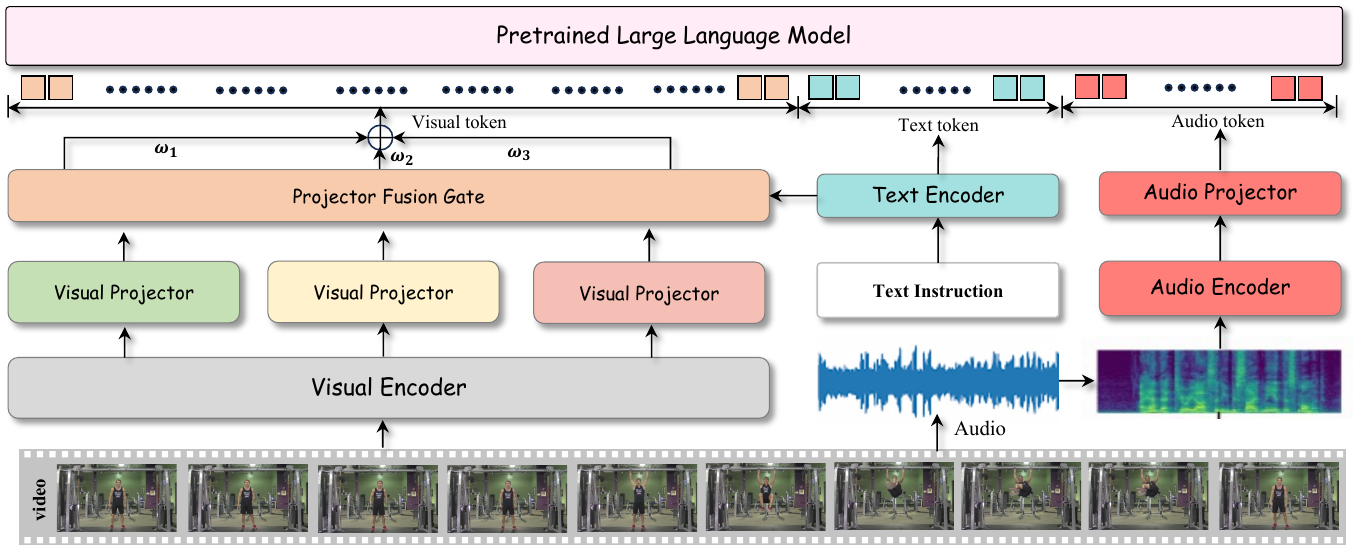}
    \vspace{-0.3cm}
    \caption{\textbf{Pipeline of HumanOmni.} 
    HumanOmni is a vision-speech language model that focus on human-centric scenes. 
    For the visual component, we pre-trained three distinct branches using separate data. The features from these branches are fused based on user instructions. HumanOmni also supports audio input, enhancing its ability to fully understand complex human-centric scenes.
    \vspace{-0.5cm}}
    \label{fig: pipeline} 
\end{figure*}

In Fig. \ref{fig: pipeline}, we illustrate the HumanOmni pipeline, which is capable of processing a multimodal input encompassing textual, auditory and visual data.

For visual component, facial expressions, body movements, individual attributes, and interactions with the environment are crucial elements for understanding human-centric video content. Different types of features are critical for different tasks; for instance, emotion analysis heavily relies on facial expressions, action recognition focuses more on body movements, and social interaction analysis depends on interactions between individuals and their environment or objects. To address these diverse requirements, we have designed three specialized branches: the Face-related Branch, Body-related Branch, and Interaction-related Branch. These branches capture distinct features to enhance the model's performance across various human-related tasks. 
Leveraging advanced visual encoders SigLIP~\cite{zhai2023sigmoid} and large language models Qwen2.5~\cite{qwen2.5}, which exhibit strong feature extraction and representation capabilities, our branch architectures remain flexible and do not require task-specific modifications. To guide each branch to focus on specific tasks, we train them using different video clips and instructions, ensuring that each branch specializes in extracting different types of features, as detailed in the training section.

In particular, while the three branches share a generic architecture, they differ in their visual projector components. 
The face-related branch employs a detail-sensitive projector MLP2xGeLU~\cite{li2024llavaone} to better capture subtle facial changes. In contrast, the body-related branch and interaction-related branch utilize a spatial-temporal projector STC~\cite{damonlpsg2024videollama2}, handling continuous actions and interaction scenes. 
Importantly, despite using two different types of projectors, the features derived from both approaches remain spatially and temporally aligned, ensuring consistency and effectiveness in feature fusion.

The features from these three branches are complementary to some extent. However, directly concatenating them would lead to an excessive number of visual tokens, imposing additional computational and analytical burdens on the LLM. While simply summing the features is one approach, we have devised a more sophisticated method for feature fusion. Inspired by LLAVA-Octopus~\cite{zhao2025llavaoctopusunlockinginstructiondrivenadaptive}, we use the rich information contained in the user instructions to dynamically adjust the weighting of features from each branch. For example, when the instruction pertains to emotion recognition, the model places greater emphasis on features from the face-related branch; for interaction scenes, it prioritizes the interaction-related branch.

Specifically, to process user instructions, we employ BERT~\cite{devlin2018bert} for encoding the commands. We focus on the [CLS] token produced by BERT, which encapsulates the semantic essence of the instruction. We chose BERT as our text encoder due to its robust pre-training, enabling it to capture deep semantic information from text.
%
BERT utilizes a bidirectional transformer architecture to encode input text, with the [CLS] token effectively summarizing the semantics of the entire sentence. This provides a strong foundation for subsequent weight generation processes.

Next, we introduce two MLPs for generating feature weights. The first MLP receives the [CLS] token as input and, through multiple layers of neural network processing, produces intermediate feature representations that capture high-level semantic details from the instructions.
%
The second MLP then takes these intermediate representations as input and further refines them to generate final weight values, each corresponding to one of the visual projectors. These generated weights are used to dynamically adjust and combine the visual features extracted by the three projectors, selecting the most suitable features for the task at hand.
%
Suppose the three projectors extract features denoted as $F_1$, $F_2$ and $F_3$, and the generated weights are $w_1$, $w_2$ and $w_3$, respectively. Then, the final visual representation $F$ is given by:
\begin{equation}
    F = w_1 \cdot F_1 + w_2 \cdot F_2 + w_3 \cdot F_3.
    \label{eq:feature_combination}
\end{equation}

This instruction-driven feature fusion approach enhances the model's flexibility and adaptability while ensuring efficient resource utilization. It allows the model to automatically adjust its focus on different types of features based on task requirements. 

For the auditory component, we follow \cite{yang2025omni} utilizing the audio preprocessor and encoder from Whisper-large-v3~\cite{radford2022whisper} to process audio data. Specifically, the audio input first undergoes preliminary processing through the audio preprocessor, generating a format suitable for encoding. Subsequently, the preprocessed audio data is encoded using Whisper's encoder, extracting robust audio features.

To ensure that audio features can be effectively integrated with visual and textual features in the same domain, we employ MLP2xGeLU as the projector. This projector maps the audio features into the text domain. 

For the text, we directly used the corresponding text encoder module from the LLM to encode the text.  Consequently, the audio tokens, along with visual and text tokens, are concatenated within a unified representation space using specific tokens to distinguish between features from different modalities, and then fed into the LLM decoder for further processing.



\section{Human-Centric Data Collection}

Although there are currently many multimodal annotated datasets, including OCR and visual navigation, there is a lack of a large-scale human-centric dataset with fine-grained annotations, limiting the development of human-centric video understanding.
Based on the existing large-scale video datasets, we have carefully designed a data processing workflow and present the largest human-centric dataset for comprehensive human-centric video understanding.

\begin{figure*}[t]
    \vspace{-5mm}
    \centering
    \includegraphics[width=1\linewidth]{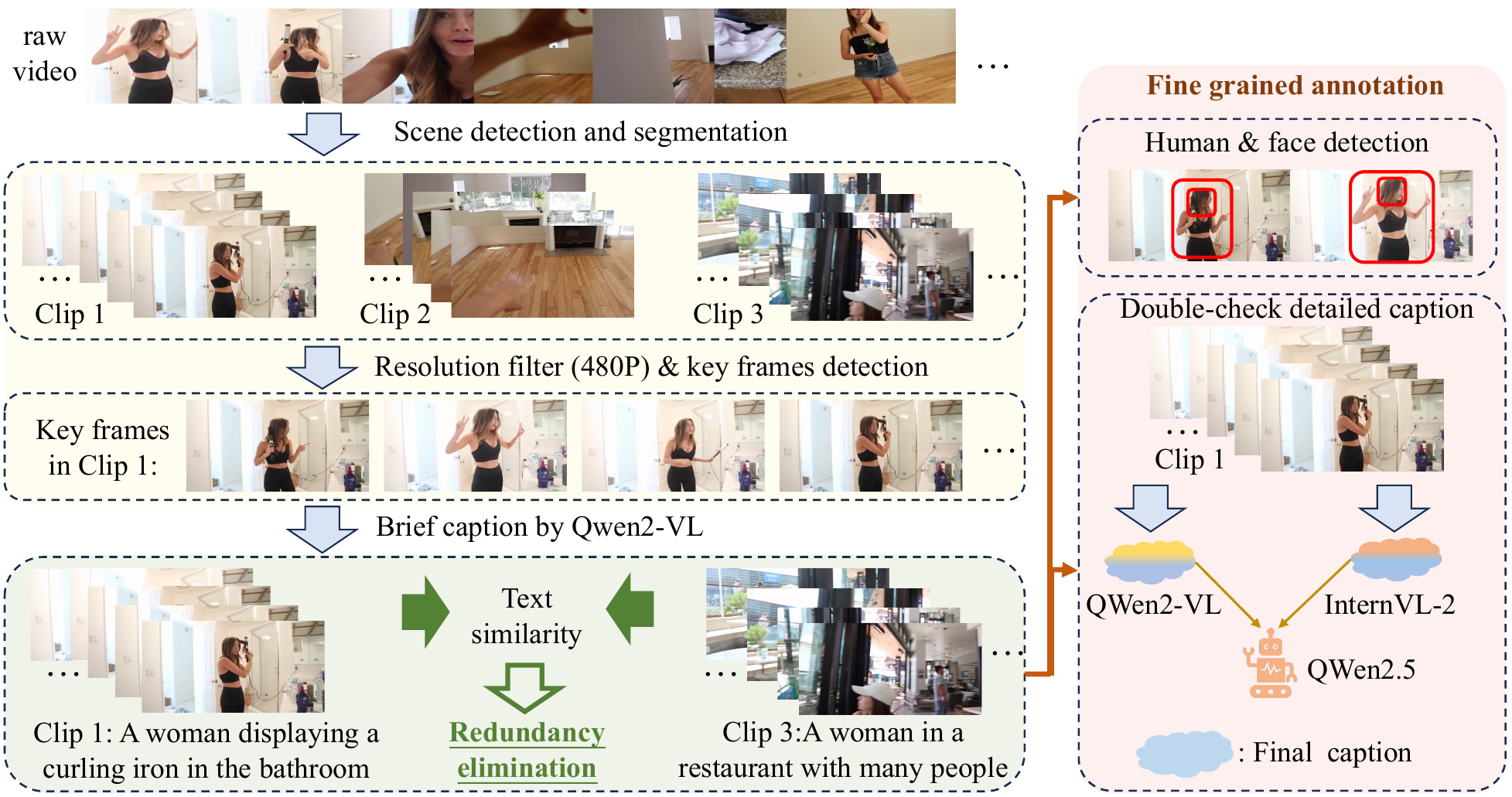}
    \vspace{-0.6cm}
    \caption{\textbf{Data processing flow.} We employ scene detection and segmentation to divide the video into clips to prevent unnatural temporal changes caused by instantaneous scene transitions. Then, the clips with relatively low resolution are removed, and the key frames detection algorithms are applied, which helps to quantify the temporal changes in clips. To further improve learning efficiency, we generate brief captions based on advanced multimodal model, and eliminate the clips similar in contexts. Finally, in addition to being automatically annotated with human and face bounding boxes, the remaining video clips will be processed by several state-of-the-art multimodal models to generate detailed captions. Subsequently, a large language model will be used to synthesize the common content across these captions, while filtering out unique content that may result from model hallucinations. 
    \vspace{-0.5cm}}
    \label{fig: data_process} 
\end{figure*}

\begin{figure*}[t]
    \vspace{-5mm}
    \centering
    \includegraphics[width=1\linewidth]{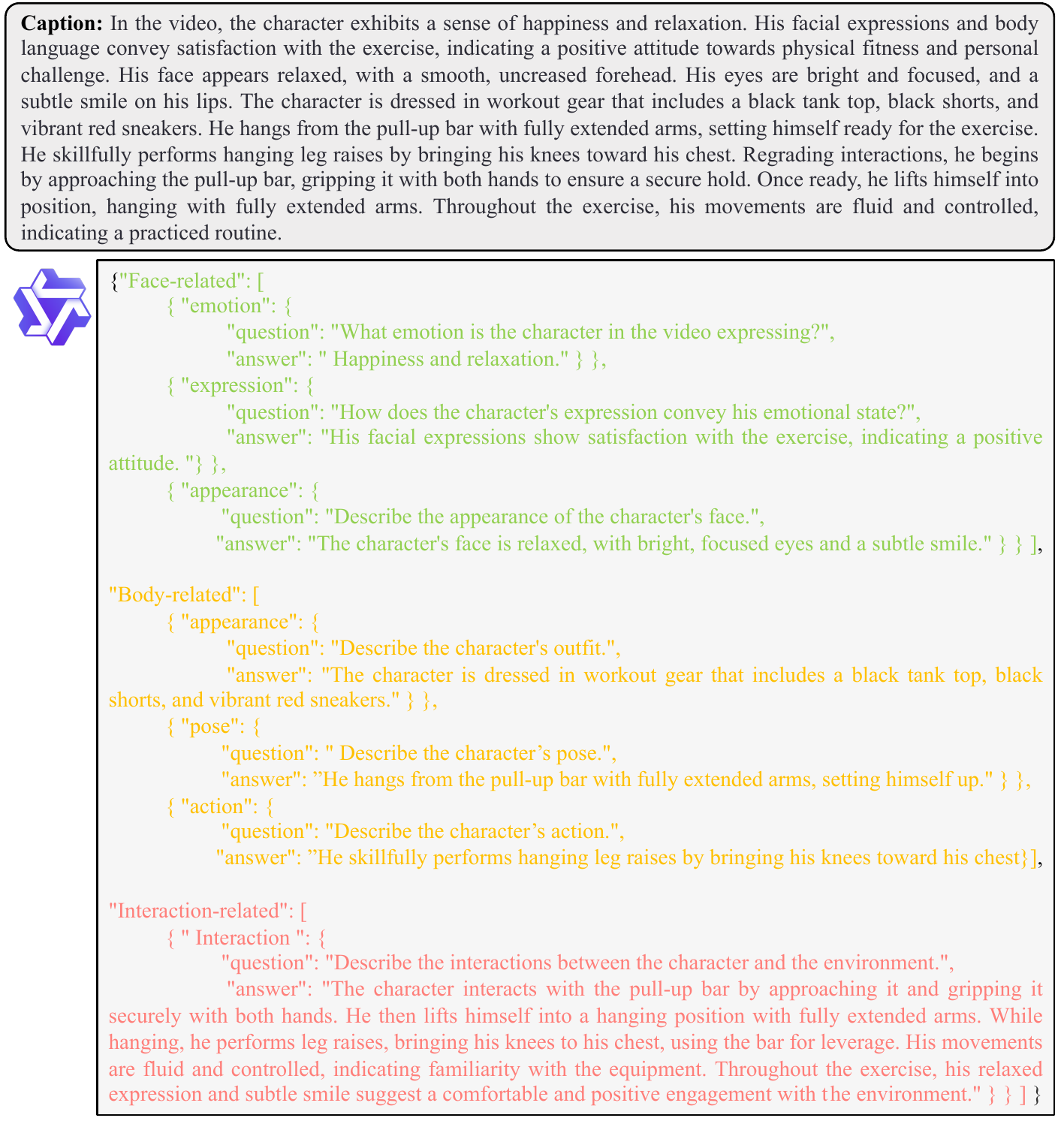}
    \vspace{-0.3cm}
    \caption{Instruction Data Generation Process for face-related, body-related, and interaction-related branches. We generate structured instruction data by leveraging Qwen2.5 with specifically designed prompts to process the detailed captions we have previously obtained. 
    \vspace{-0.5cm}}
    \label{fig: pipeline_2} 
\end{figure*} 

\begin{figure*}[t]
    \vspace{-5mm}
    \centering
    \includegraphics[width=1\linewidth]{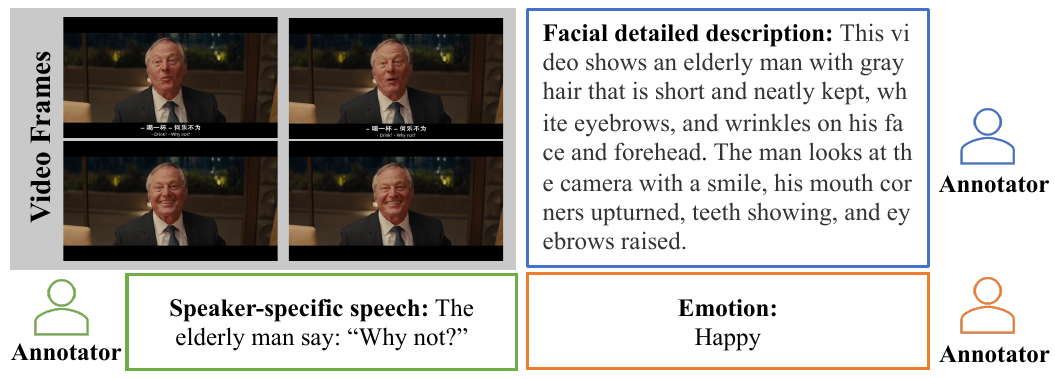}
    \vspace{-0.3cm}
    \caption{Illustration of the data annotation process. We annotate the data from the perspectives of Emotion, Speaker-specific speech, and Facial detailed description.
}
    \vspace{-0.5cm}
    \label{fig: pipeline_3} 
\end{figure*} 

\subsection{Video Collection}
We form our dataset based on the existing web-scale dataset: Panda-70M~\cite{chen2024panda70m} which covers various scenes and contents. 
Despite initial processing and caption generation, much useful information, particularly related to human subjects, remains underutilized. Existing research has demonstrated that data quality is crucial for model performance; simply increasing the quantity of low-quality data does not lead to significant performance improvements. Therefore, we have further optimized our dataset to create a human-centric collection with the following characteristics: high video resolution (above 480P), rich temporal dynamics, inclusion of face and body detection bounding boxes, and captions verified through a dual-check process.
Our data processing includes the following step and an illustration is in Fig.~\ref{fig: data_process}.

\textbf{ - Stage 1: Temporal processing.} Temporal dynamics is a key feature distinguishing video data from static images. Videos lacking temporal changes offer limited learning value, while those with excessive and unrealistic temporal variations can confuse models.
To extract naturally smooth temporal changes in the video, we employed scene recognition algorithms to identify multiple scenes within videos and segment them into clips, as shown in the center of Fig.~\ref{fig: data_process}. Scene detection and segmentation avoid drastic changes from multiple different perspectives in the video clip. Besides, it ensures that clips with intense temporal changes are broken down into shorter segments, which can then be filtered out. 

Furthermore, we extracted keyframes for each clip to find the temporal variations. The clips with minimal temporal changes would have very few keyframes identified, allowing us to screen out such videos based on the number of detected keyframes. We also removed clips with resolutions below 480P to enhance the overall quality of the data.

\textbf{ - Stage 2: Reducing Redundancy.}
To improve learning efficiency, we eliminated redundant video segments with similar contexts or meanings, reducing data redundancy as shown in the bottom of Fig.~\ref{fig: data_process}.
Specifically, we used an advanced multimodal model, QWen2-VL-72B~\cite{Qwen2-VL}, to generate brief descriptions for each clip, focusing on the general contextual information. By calculating the semantic similarity between these brief descriptions using language model, we were able to filter out clips with high semantic similarity and repeated patterns. 

\textbf{ - Stage 3: Fine-grained Annotations.}
For the remaining data, we generated detailed descriptions using the advanced multimodal model (QWen2-VL-72B) to maximize the utility of the data as shown on the right side of Fig.~\ref{fig: data_process}.
Given the well-known issue of hallucination in large models, we implemented a dual verification method to eliminate hallucinations from detailed captions. 
Specifically, we used an additional multimodal large model~\cite{wang2024internvideo2} to generate multiple detailed descriptions for each clip. Given that the content of hallucinations is not related to the actual content of in the video, we hypothesize that hallucinations generated by different models are distinct. Hence, we heuristically employed a large language model (QWen2.5-72B~\cite{qwen2.5}) to summarize the common points across the different detailed descriptions, ensuring the accuracy of the final descriptions and effectively removing hallucinations.
Finally, we apply advanced detection algorithms to detect the persons and the faces in the video clip, providing rich face and body detection bounding boxes. These bounding boxes enable fine-grained alignment between the human-centric visual content and other annotations (e.g., detailed caption).

Through the aforementioned steps, we collected 2.4M human-centric video clips with captions.
We then utilize Qwen2.5-72B to generate structured data from detail captions for the pre-training of different branches, as illustrated in the Fig.~\ref{fig: pipeline_2}. 
We systematically constructed instruction pairs for the face branch, body branch, and interaction branch by utilizing the structured data. 

\subsection{Instructions Generation}

For the face-related branch, we filtered out videos that did not include descriptions of faces, emotions, or expressions. This resulted in a dataset of 1.78M videos. We then created detailed instruction pairs for facial features, emotions, and expressions, totaling 4.12M instruction pairs. These pairs were used to train the face-related branch.

For the body-related branch, we applied a similar method. We filtered out videos that lacked information on human appearance attributes, actions, or poses, leaving us with 2.21M videos. We then created specific instruction pairs for human appearance attributes, actions, and poses, resulting in a total of 5.75M instruction pairs. These pairs were used to train the body-related branch.

Finally, for the interaction-related module, we created specific instruction pairs for interactions with the external environment. Additionally, to ensure that our caption information was fully utilized, we incorporated detailed captions as instruction data in this module. This process resulted in a total of 4.8M instruction pairs, including 2.4M interaction instruction pairs and 2.4M detailed caption instruction pairs. These pairs were used to train the interaction-related module.

All of these instructions were used during the pre-training phase. The instructions come from captions that we double-checked, ensuring higher accuracy. By reasonably segmenting and filtering video clips, we also improved video quality, which helps the model better understand human-centric scenes.

Additionally, we manually annotated a subset of our human-centric video data by randomly sampling 50K video clips containing both vision and speech. These clips were annotated for emotion recognition, detailed facial expression descriptions, and speaker-specific speech annotations. Due to varying annotation difficulties, we completed emotion annotations for all 50K videos, detailed facial expression descriptions for 5K videos, and speaker-specific speech annotations for 20K videos, resulting in a total of 75K instruction pairs.
The process is illustrated in Fig.~\ref{fig: pipeline_3}.
These annotated data were used in the fine-tuning of the visual component, further enhancing its feature extraction capabilities. 
Because these data include both visual and speech components and are manually annotated for high quality, they were also utilized in Cross-Modal Interaction Integration.

\section{Training}
To build a multimodal large model capable of accurately understanding human-centric video information and possessing cross-modal interaction capabilities, our training strategy is divided into three stages. Initially, we focus on pretraining and fine-tuning the model's visual abilities using a substantial amount of human-centric video data, enabling the model to learn rich spatio-temporal feature representations and patterns of human behavior, thereby achieving a deep understanding of human-related details in video content. Next, we conduct standalone audio capability training with audio data, allowing the model to recognize and interpret speech. Finally, we perform cross-modal interaction training by integrating auditory and visual data, enhancing the model's ability to process and associate information across different modalities, ensuring it can provide accurate understanding and responses in complex multimedia environments. 

\subsection{Visual Capability Construction}
Our model's visual component includes three specialized branches: face-related branch, body-related branch, and interaction-related branch. For each branch, we generated specific instruction data as described in the above section. This instruction data was used to pre-train each branch, during which only the projector parameters were updated. The aim was to keep all other parameters identical across the three branches to facilitate integration during fine-tuning.

During fine-tuning, we used manually annotated data consisting of 50K emotion recognition instructions and 5K facial expression description instructions, along with general Oryx~\cite{liu2024oryx} fine-tuning data. We integrated the three branches using an instruction-driven fusion module. In this process, we froze the parameters of the visual encoder and BERT, while training the parameters of the three projectors, the large language model, and two MLPs that generate the fusion weights.

During this phase, even though some of the videos contain both auditory and visual information, we only utilized the visual part.

\subsection{Auditory Capability Development}
In this stage, we aim to align the modalities between text and audio, enhancing the model's ability to understand and respond to audio in various contexts. We exclusively sample audio data from tasks such as automatic audio captioning, automatic speech recognition, and sound event classification, resulting in a total of approximately 18,000 hours of data used to train the audio projector.

Specifically, we utilize the WavCaps~\cite{mei2024wavcaps} dataset, which provides around 7,500 hours of annotated audio, offering detailed captions that describe the audio events. This dataset plays a crucial role in helping the model understand and generate descriptive audio analyses.
We also select multiple comprehensive ASR datasets including WenetSpeech~\cite{zhang2022wenetspeech}, GigaSpeech~\cite{chen2021gigaspeech}, CommonVoice15~\cite{commonvoice:2020}, and LibriSpeech~\cite{LibriSpeech}. These datasets cover extensive and diverse speech data, which are important in training models for speech recognition tasks. 
For SEC, the VGGSound~\cite{chen2020vggsound} dataset is chosen due to its extensive collection of audio events.
For the different tasks, we designed multiple question templates to prompt the model in generating captions, performing speech recognition, and classifying sounds, which in turn enables the model to thoroughly understand and process human-related audio information.
\begin{table}[ht]
    \centering
    \begin{tabular}{>{\raggedright\arraybackslash}p{2cm}|>{\raggedright\arraybackslash}p{6cm}|>{\raggedright\arraybackslash}p{3cm}}
        \hline
        \textbf{Task Type} & \textbf{Datasets} & \textbf{Duration (hours)} \\
        \hline
        AAC & WavCaps~\cite{mei2024wavcaps} & \textasciitilde 7.5k\\ 
        \hline
        \multirow{2}{*}{ASR} & WenetSpeech~\cite{zhang2022wenetspeech}, GigaSpeech~\cite{chen2021gigaspeech}, CommonVoice15~\cite{commonvoice:2020}, LibriSpeech~\cite{LibriSpeech} & \multirow{2}{*}{\textasciitilde 10k} \\
        \hline
        SEC & VGGSound~\cite{chen2020vggsound} & \textasciitilde 0.5k \\
        \hline
    \end{tabular}
    \caption{Details of audio datasets for training audio projectors.}
    \label{tab:data_sets}
\end{table}

We use the encoder and the audio preprocessor from the Whisper-large-v3~\cite{radford2022whisper} as the audio encoder and processor. Specifically, We resample each audio to a frequency of 16kHz and convert the waveform into 128-channel mel-spectrogram using a window size of 25ms and a hop size of 10ms.
To reduce the token length of the audio, we introduce an average pooling layer with a stride of 3, resulting in each audio frame from the audio encoder corresponding to a 60ms segment of the original audio. We use two linear layers to connect this to the LLM decoder. Additionally, we wrap each audio embedding with a pair of special tokens to indicate the start and end positions of the audio embedding.
\subsection{Cross-Modal Interaction Integration}
To enhance our model's video-audio interaction capabilities, we synthesized a series of visual-auditory cross-modal interaction data. For audio data, we collected a diverse dataset covering various audio tasks, including samples from the audio pre-training phase, emotion recognition datasets, and audio question-answering datasets, totaling 7,000 hours of audio. 
For video data, we used all the aforementioned manually annotated 20K speaker-specific speech recognition data, as well as the instruction data used for visual fine-tuning. Additionally, we incorporate multi-modal emotion recognition datasets, converting classification labels with GPT-4o into a question-answer format, which includes DFEW~\cite{jiang2020dfew}, MAFW~\cite{liu_mafw_2022}, CAER~\cite{lee2019context}, and FERV39k~\cite{wang2022ferv39klargescalemultiscenedataset}. 

To better distinguish features from different modalities, we encapsulate the embeddings of audio and visual data using distinct special tokens. We initialize the visual projectors and LLM decoder with parameters obtained from the Visual Capability Construction phase, while the audio projector is initialized with parameters from the Auditory Capability Development phase. During this training phase, we jointly fine-tune the LLM decoder, all projectors and two multi-layer perceptrons (MLPs) that generate the fusion weights to optimize their performance in handling multi-modal inputs.

To ensure that our HumanOmni can understand both scenes that include visual and auditory information and those with only visual input, for each video that contains audio, we also generate a version without audio for training. The model determines which modality to use based on special tokens in the instructions.
Additionally, if either the auditory or visual part is missing, we fill in with default tokens to ensure consistent and complete inputs. This design allows the model to maintain stable performance across different modality combinations.

\section{Experiments}
We evaluated HumanOmni's ability to understand audio-visual inputs on several human-related tasks, such as emotion recognition, facial expression description, and action understanding. We also tested HumanOmni's performance on speech recognition using only audio inputs. Finally, we explored how different modalities affect model performance across these human-centric tasks.

\begin{table*}[t]
    \centering
    \definecolor{lightlightgray}{gray}{0.8}
    \tablestyle{12pt}{1.1}
    \begin{tabular}{lccccc}
        \toprule
        \multirow{2}{*}{\textbf{Method}} &\multirow{2}{*}{\textbf{Modalities}}   & \multicolumn{2}{c}{\textbf{DFEW}} & \multicolumn{2}{c}{\textbf{MAFW}} \\
        \cmidrule(lr){3-4} \cmidrule(lr){5-6} 
       & & \textbf{UAR} & \textbf{WAR} & \textbf{UAR} & \textbf{WAR} \\
        \midrule
       
                \rowcolor[HTML]{EFEFEF} \multicolumn{6}{c}{\textbf{Specialized models for emotion-related tasks}}\\
        Wav2Vec2.0~\cite{baevski2020wav2vec}& A& 36.15 & 43.05 & 21.59 & 29.69\\
        HuBERT~\cite{soft-vc-2022} & A & 35.98 & 43.24 & 25.00 & 32.60\\
        DFER-CLIP~\cite{zhao2023dferclip} & V  & 59.61 & 71.25 & 38.89 & 52.55 \\
        MAE-DFER~\cite{sun2023mae}& V & 63.41 & 74.43 & 41.62 & 54.31\\
        HiCMAE~\cite{sun2024hicmae} & AV& 63.76 & 75.01 & 42.65 & 56.17\\
        Emotion-LLaMA~\cite{cheng2024emotion}& AV & 64.21 & 77.06 & - & - \\
        MMA-DFER & AV& 66.85 & 77.43 & 44.25 & 58.45\\
        
        \midrule

       
        
         Qwen2-VL-7B~\cite{Qwen2-VL}& V & 43.08 & 52.83 & 31.67 & 45.89\\
        Qwen2-VL-72B~\cite{Qwen2-VL} & V& 39.24 & 45.12 & 42.61 & 46.07\\
     VITA~\cite{fu2024vitaopensourceinteractiveomni}& AV& 21.36 & 32.07 & 14.05 & 33.38 \\
     InternLM-XComposer-2.5-OL~\cite{internlmxcomposer2_5_OL}& AV& 44.23 & 51.29 & 33.78 & 46.81  \\
     GPT4-O~\cite{gpt4v}   & AV& 50.57 & 57.19 & 38.29 & 48.82 \\
        \rowcolor[HTML]{EFEFEF} \textbf{\MyMthd{}}
        & AV &\textbf{74.86} & \textbf{82.46} & \textbf{52.94} & \textbf{68.40}  \\    
        \bottomrule
    \end{tabular}
    \caption{Results on DFEW and MAFW.}
    \label{tab: human benchmark}
\end{table*}

\subsection{Evaluations on Emotion Recognition}
Both DFEW and MAFW are video-clip-based datasets designed for Dynamic Facial Emotion Recognition task, with DFEW providing a 7-dimensional expression distribution vector and MAFW providing an 11-dimensional expression distribution vector for each video clip.

As shown in \tabref{tab: human benchmark}, while VLM methods possess broader capabilities, they still exhibit a performance gap compared to specialized methods in dynamic emotion recognition tasks. In this task, both video and audio information play crucial roles, which is where the HumanOmni model excels. Experimental results demonstrate that HumanOmni significantly outperforms existing video-language multimodal  models, audio-language multimodal large models, 
recently proposed omni model and specialized methods in this field. Moreover, it also shows a clear advantage over recently proposed Omni models for emotion recognition.





\definecolor{lightlightgray}{gray}{0.95}
\begin{table*}[t]
    \centering
    \small
    \setlength{\tabcolsep}{2.8pt} 
    \renewcommand{\arraystretch}{1.2} 
    \begin{tabular}{lccccccc}
        \toprule
        \textbf{Method} &\textbf{Correctness} & \textbf{Detail} & \textbf{Context} & \textbf{Temporal} & \textbf{CIDEr} & \textbf{Rouge-L}& \textbf{AutoDQ}  \\
        \midrule
        \rowcolor[HTML]{EFEFEF}\multicolumn{8}{c}{Vision large language model}\\
        VideoLLaMA \cite{damonlpsg2023videollama}  & 3.60 & 3.67 & 3.84 & 3.50 & 0.189 & {0.196} & 0.303  \\
        VideoChat \cite{li2023videochat}  & 3.47 & 3.52 & 3.92 & 3.38 & 0.251  &  {0.192} & 0.344  \\
        VideoChat2 \cite{li2023videochat}  & 3.70 & 3.56 & 4.16 & 3.52 & 0.202 & {0.229}  & 0.311 \\
        Chat-UniVI \cite{jin2023chatunivi} & 3.64 & 3.63 & 4.21 & 3.61 & 0.189 & {0.231}  & 0.396 \\
        LLaVA-Next-Video \cite{zhang2024llavanext-video} & 4.19 & 4.07 & 4.39 &  4.04 & 0.250  & {0.249}  & 0.395\\
        ShareGPT4Video \cite{chen2024sharegpt4video}  & 4.24 & 4.13 & 4.35 &  4.09 & 0.192 & {0.205} & 0.394  \\
        LLaMA-VID \cite{li2024llamavid} & 3.95 & 4.01 & 4.22 & 3.71 & 0.195 & {0.231} & 0.339  \\
        VideoLLaMA2 \cite{damonlpsg2024videollama2}  & 4.17 & 4.02 & 4.47 & 3.93 & 0.253& {0.266} & 0.344  \\
        PLLaVA \cite{xu2024pllava}  & 4.21 & 4.15 & 4.37 & 4.08 & 0.268 & {0.250} & 0.393  \\
        ST-LLM \cite{liu2024st}  & 4.00 & 3.98 & 4.31 & 3.94 & 0.213& {0.238} & 0.321  \\
        Tarsier \cite{wang2024tarsierrecipestrainingevaluating}  & 3.59 & 3.50 & 4.07 & 3.41 & 0.143 & {0.185} & 0.415  \\
        LLaVA-OneVision \cite{li2024llavaone} & 3.68 & 3.47 & 4.10 & 3.42 & 0.115 & {0.165} & 0.379  \\
        {FaceTrack-MM~\cite{zhao2025facialdynamicsvideoinstruction}}  &{4.42} & {4.30} & {4.60} &{4.26} & \textbf{0.418} & \textbf{0.473} & {0.483} \\
        Qwen2-VL-72B \cite{Qwen2-VL}  & 4.28 & 4.14 & 4.55 & 4.08 & 0.241 & {0.314}  & 0.449 \\
        Qwen2-VL-7B \cite{Qwen2-VL}  & 4.23 & 4.16 & 4.52 & 4.02 & 0.204 & {0.233} & 0.422  \\
        Qwen2-VL-2B \cite{Qwen2-VL}  & 4.01 & 3.98 & 4.37 & 3.88 & 0.202 & {0.221}  & 0.406 \\
        Claude3.5-Sonnet \cite{Claude2024} & 4.13 & 4.01 & 4.49 & 4.05 & 0.243 & {0.228} & 0.442  \\
            \midrule
            \rowcolor[HTML]{EFEFEF}\multicolumn{8}{c}{Omni-modality large language model}\\
             GPT4-O \cite{openai2024gpt4o} & 4.22 & 3.97 & 4.48 & 3.90 & 0.264 & {0.213} & 0.432  \\
     VITA~\cite{fu2024vitaopensourceinteractiveomni}& 3.98 & 3.74 & 4.11 & 3.59  & 0.191 & 0.224 & 0.366 \\
     InternLM-XComposer-2.5-OL~\cite{internlmxcomposer2_5_OL}& 3.91 & 3.70 & 4.12 & 3.54  & 0.113 & 0.164 & 0.382 \\
        \rowcolor[HTML]{EFEFEF} \textbf{HumanOmni}  & \textbf{4.58} & \textbf{4.41} & \textbf{4.70} & \textbf{4.41} & {0.412}  & {0.468} & \textbf{0.523} \\
        \bottomrule
    \end{tabular}
    \caption{Comparison of different methods on DFEC~\cite{zhao2025facialdynamicsvideoinstruction}  benchmark.}
    \vspace{-5pt}
    \label{tab:dfec}

  \end{table*}

\subsection{Evaluations on Facial Expression}

Facial expressions refer to external features displayed through facial muscle movements, such as smiling or frowning, while emotions denote internal emotional states, such as happiness or sadness. Although facial expressions are one way to convey emotions, not all expressions directly correspond to specific emotional states, and the same expression can represent different emotions in various contexts.
In this evaluation, we utilized the recently proposed DFEC dataset for facial expression description and adopted the evaluation methods recommended by DFEC. 

In Tab.~\ref{tab:dfec}, our experimental results show that the HumanOmni model with combined video and audio input not only outperforms other open-source models but also surpasses the FaceTrack-MM~\cite{zhao2025facialdynamicsvideoinstruction} method proposed in DFEC, achieving superior performance in facial expression description tasks.

\subsection{Evaluations on Actions Understanding}
MVBench is a comprehensive video understanding benchmark covering 20 tasks organized in the form of multiple-choice questions. 
From this extensive suite of challenges, we select a specialized benchmark focusing on human-related subtasks, demonstrating in \tabref{tab: mvbench_filtered_with_avg}. Specifically, we have selected six tasks most pertinent to human behavior analysis: Action Sequence (AS), Action Antonym (AA), Unexpected Action (UA), Object Interaction (OI), Action Count (AC), and Fine-grained Action (FA).
This refined selection aims to provide a focused evaluation framework for the nuanced aspects of human activity recognition within video content.

\begin{table}[t]
  \centering
  \tablestyle{8.5pt}{1.1}
  \begin{tabular}{lccccccl} \toprule
    \textbf{Method}  & \textbf{AS} & \textbf{UA} & \textbf{AA} & \textbf{OI} & \textbf{AC} & \textbf{FA} & \textbf{Avg} \\ 
    \midrule 
     
    \rowcolor[HTML]{EFEFEF}\multicolumn{8}{c}{Vision large language model}\\
    Otter-V~\cite{li2023otter} & 23.0 & 29.5 & 27.5 & 28.0  & 26.0 & 27.0 & 26.8 \\
     mPLUG-Owl-V~\cite{ye2023mplugowl}& 22.0  & 29.0 & 34.0 & 27.0  & 31.5 & 29.0 & 28.8 \\
     Video-LLaMA~\cite{damonlpsg2023videollama}  
     & 27.5 & 39.0 & 51.0 & 40.5 & 34.0 & 29.0 & 36.8 \\
     LLaMA-Adapter~\cite{zhang2023llamaadapter} 
     & 23.0 & 33.0 & 51.0 & 32.5  & 29.0 & 30.0 & 33.1 \\
     Video-ChatGPT~\cite{Maaz2023VideoChatGPT} 
     & 23.5 & 26.5 & 62.0 & 28.0  & 30.5 & 22.5 & 32.2 \\
     VideoChat~\cite{li2023videochat} 
     & 33.5 & 40.5 & 56.0 & 40.5  & 35.0 & 33.5 & 39.8 \\
     VideoChat2~\cite{li2024mvbenchcomprehensivemultimodalvideo}
     & 75.5 & 60.5 & 83.5 & 74.5  & 37.0 & 50.5 & 63.6 \\
     ST-LLM~\cite{liu2024st}
     & 66.0 & 58.5 & 84.0 & 73.5  & 36.5 & 44.0 & 60.4 \\
     PLLaVA~\cite{xu2024pllava}
     & 58.0 & 61.0 & 55.5 & 61.0 & 39.5 & 41.0 & 52.6 \\
     VideoLLaMB~\cite{videollamb}
     & 54.5 & 52.0 & 86.5 & 58.5  & 40.5 & 44.5 & 56.1 \\
     Qwen2-VL-72B*~\cite{Qwen2-VL}& 51.5 & 82.0 & 93.5 & 81.5  & 48.5 & 49.0 & 67.7  \\
     Qwen2-VL-7B*~\cite{Qwen2-VL}& 73.5 & 80.0 & 79.0 & 78.5  & 46.0 & 49.0 & 67.7 \\
     Qwen2-VL-2B*~\cite{Qwen2-VL}& 77.5 & 76.5 & 76.5 & 77.5  & 50.0 &  47.5 & 67.6 \\     
     GPT-4V~\cite{gpt4v}& 55.5 & 63.5 & 72.0 & 59.0  & 39.0 & 47.5 & 56.1\\ 
    \midrule
    \rowcolor[HTML]{EFEFEF}\multicolumn{8}{c}{Omni-modality large language model}\\
     VITA~\cite{fu2024vitaopensourceinteractiveomni}& 58.0 & 81.5 & 73.5 & 61.5  & 45.5 & 42.0 & 60.3 \\
     InternLM-XComposer-2.5-OL~\cite{internlmxcomposer2_5_OL}& 84.5 & 81.0 & 75.0 & 79.5  & 60.5 &  46.0 & 71.1 \\
    
    \rowcolor[HTML]{EFEFEF} \textbf{\MyMthd{}} & {70.0} & {78.0}  & {92.5}  & {80.5}  & {65.5}  & {49.0}  & \textbf{72.6}  \\
    \bottomrule
  \end{tabular}
  \vspace{3pt}
    \caption{Results on MVBench. We select a subset of human-related subtasks from MVbench.}
    \vspace{-15pt}
  \label{tab: mvbench_filtered_with_avg}
\end{table}

The experimental results show that on the MVBench dataset, HumanOmni significantly outperforms nearly all mainstream methods with the same parameter size, with the exception of a few methods that utilized the full MVBench dataset. 

\subsection{Evaluations on Speech Recognition}

Speech recognition capability is a crucial component of human-computer interaction. To demonstrate the advantages of our approach within the domain of speech recognition, \tabref{tab: audio benchmark} presents results from four widely recognized benchmarks in this field: LibriSpeech~\cite{LibriSpeech}, WenetSpeech~\cite{zhang2022wenetspeech},  and Fleurs~\cite{fleurs2022arxiv}. These benchmarks are specifically chosen for their distinct characteristics and contributions to evaluating speech recognition systems across different languages and contexts. LibriSpeech focuses on English speech recognition, while Fleurs is dedicated to evaluating cross-lingual speech representations. 
From the table, it can be seen that our method is leading among the current Omni models. However, compared to proprietary speech recognition approaches, current audio-visual methods can still be improved.
\begin{table}[h]
    \centering
    \definecolor{lightlightgray}{gray}{1}
     \setlength{\tabcolsep}{14pt} 
    \begin{tabular}{lccc}
        \toprule
        \multirow{3}{*}{\textbf{Method}} &   \multirow{3}{*}{\begin{tabular}[c]{@{}c@{}}\textbf{Librispeech}\\      \textit{dev-clean} | \textit{dev-other} | \\ \textit{test-clean} | \textit{test-other} \end{tabular}}  & \multirow{3}{*}{\begin{tabular}[c]{@{}c@{}}\textbf{WenetSpeech}\\      \textit{test-net} \\| \textit{test-meeting} \end{tabular}}  &\multirow{3}{*}{\begin{tabular}[c]{@{}c@{}}\textbf{Fleurs}\\      \textit{en} | \textit{zh}  \\ {\quad}\end{tabular}}\\
        \\
        \\
        \midrule
  
          \rowcolor[HTML]{EFEFEF}\multicolumn{4}{c}{Audio large language model}\\
        Qwen2-Audio~\cite{Qwen2-Audio} & \textbf{1.3}|\textbf{3.4}|\textbf{1.6}|\textbf{3.6} &7.8|8.4 & 9.0|15.7 \\
        SenseVoice-L~\cite{an2024funaudiollm} & - | - |2.5|4.2 &\textbf{6.0}|\textbf{6.7} & - \\
      \midrule
         \rowcolor[HTML]{EFEFEF}\multicolumn{4}{c}{Omni-modality large language model}\\
        Baichuan-Omni~\cite{li2024baichuanomni} & - &6.9|8.4  & 7.0|\textbf{4.7} \\
        VITA~\cite{fu2024vitaopensourceinteractiveomni} & 7.6|16.6|8.1|18.4 & 12.2|16.5  & - \\
        Mini-Omni2~\cite{xie2024miniomni2opensourcegpt4ovision} & 4.7|9.4|4.8|9.8 & -& - \\
        \rowcolor[HTML]{EFEFEF} \textbf{HumanOmni}
        & {3.8|7.5|3.7|8.0} & {10.4|8.4} & {\textbf{6.3}|7.5} \\        
        \bottomrule
    \end{tabular}
    \vspace{3pt}
    \caption{Results on Audio Benchmarks.}
    \label{tab: audio benchmark}
\end{table}

\subsection{Explorations of Modality Effects}
Here we explored modalities efects on human-centric task performance.
In Table~\ref{tab:modality_comparison}, we evaluated HumanOmni's performance on emotion recognition, facial expression description, and action understanding under different input modalities. As expected, in the emotion recognition task, single-modal configurations (using only video or only audio) performed notably lower compared to the multi-modal configuration that used both video and audio inputs. For facial expression description, even when using only video input, the HumanOmni model maintained excellent performance, only slightly lower than with combined inputs. This is because facial expression recognition primarily relies on visual information, with limited added value from audio data. In the action understanding task, where actions are mainly represented by visual content, the contribution of audio was even more limited, as confirmed by our experimental results.
These results demonstrate HumanOmni's robust performance across different input modalities. Additionally, they show that for all tasks, the combined visual-auditory input consistently achieved the best results, underscoring the necessity of joint audio and video inputs in human-centric scenes.
\begin{table}[h]
    \centering
    \begin{tabular}{lcccc}
        \toprule
        \textbf{Input Modality} & \textbf{DFEW-WAR} & \textbf{MAFW-WAR} & \textbf{DFEC-AutoDQ} & \textbf{MVBench-Avg}  \\
        \midrule
        Video Only              &       70.62          &  59.58             & 0.510             & 72.3                \\
        Audio Only              & 58.63               & 51.33             & -             & -                \\
        Video-Audio   & 81.82                 & 66.50            & 0.523            & 72.6          \\
        \bottomrule
    \end{tabular}
            \vspace{5pt}
        \caption{Comparison of HumanOmni on Different Input Modalities}
\vspace{-3pt}
        \label{tab:modality_comparison}
\end{table}

\section{Conclusion}
In this work, we developed HumanOmni, the first human-centric multi-modal large language model. We constructed a dataset containing over 2.4 million human-centric video clips annotated with more than 14 million detailed captions and instructions to facilitate the understanding of diverse human-centered scenes.
HumanOmni features a specialized architecture with three branches: a face-related branch, a body-related branch, and an interaction-related branch. Each branch addresses specific categories of human-centric scenes. By using user instructions to guide the adaptive fusion of features from these branches, HumanOmni significantly enhances its robustness across various scenarios. 
Additionally, HumanOmni supports joint audio and video input, enabling a more comprehensive understanding of scenes.
We evaluated HumanOmni's performance through extensive experiments on multiple human-centric tasks, demonstrating its effectiveness in understanding complex human-centered interactions.
To promote community-driven development and further research, we will open-source our code and model.

\bibliography{main}
\bibliographystyle{plain}
\end{document}